\pdfoutput=1
\documentclass[11pt,a4paper]{article}

\usepackage[english]{babel}
\usepackage[T1]{fontenc}
\usepackage[utf8]{inputenc}

\usepackage{amsmath,amssymb,amsfonts,mathtools}

\usepackage{geometry}
\usepackage{graphicx}
\usepackage{float}
\usepackage{microtype}

\usepackage{mathptmx}

\usepackage{caption}
\usepackage[hidelinks]{hyperref}

\graphicspath{{./}}
\makeatletter
\renewcommand{\maketitle}{
\begin{flushleft}
{\LARGE \bfseries \@title \par}
\vspace{1em}
{\normalsize \@author \par}
\vspace{1em}
\end{flushleft}
}
\makeatother
\title{
From Observed Viability to Internal Predictive Approximation:\\
A Single-Subject Latent-Space Analysis of Gait Dynamics Under Occlusal Constraint
}

\author{
Jacques Raynal$^{1,*}$,
Pierre Slangen$^{2}$,
Elsa Raynal$^{3}$,
Jacques Margerit$^{4}$\\[0.5em]
{\small $^{1}$Laboratory of Bioengineering and Nanosciences (LBN), University of Montpellier, France}\\
{\small $^{2}$EuroMov Digital Health in Motion, University of Montpellier, IMT Mines Al\`es, Al\`es, France}\\
{\small $^{3}$Certified Sophrologist and Dental Assistant, Sensorimotor Practice, Montpellier, France}\\
{\small $^{4}$Emeritus Professor, University of Montpellier, France}\\[0.5em]
{\small $^{*}$Corresponding author: \texttt{raynal.cab@gmail.com}}
}

\date{}

\begin{document}

\maketitle

\begin{abstract}

Understanding adaptive biomechanical systems requires distinguishing between observable performance, static multivariate representation, longitudinal displacement, and the possibility of approximating observed representational change.

The preceding Level~3 study showed that neither an aggregated scalar score nor a static exploratory embedding uniquely resolved the occlusal observational probes. The subsequent Level~4 study therefore shifted the analysis from static representational non-identifiability to longitudinal centroid displacement within a common PCA representation. In that selected projection, OC3 showed the lowest M1--M2 centroid displacement, ONL occupied an intermediate position, and OC2.5 showed the highest displacement.

The present study introduces a fifth analytical level centered on internal predictive approximation of this observed longitudinal displacement. The term predictive is used here in a restricted methodological sense: it refers to approximation of observed M1--M2 transformations within the same single-subject dataset, not to prospective clinical prediction, patient-level forecasting, or generalization to unseen individuals.

Using an exploratory single-subject design in a participant with Parkinson's disease, gait data were recorded with instrumented insoles under six occlusal observational probes: neutral natural occlusion (ONL), wide open-mouth disengagement without dental contact (OBL), strong voluntary clenching (OSL), a nominal 2.5-degree increase in vertical dimension of occlusion in centric relation (OC2.5), a nominal 3-degree increase in vertical dimension of occlusion in centric relation (OC3), and a nominal 3-degree increase combined with mandibular protrusion and hinge-axis displacement (OC3P). Two measurement sessions were conducted eleven weeks apart, during which the participant underwent a structured sensorimotor intervention.

A common PCA representation was used to describe the observed M1--M2 transformations. A simplified feed-forward neural network was then trained to approximate these transformations directly in the selected PC1--PC2 coordinate system. Occlusal configurations were treated as observational probes applied during measurement, not as continuous causal drivers of longitudinal evolution.

Within the core Level~4-aligned analysis, the model approximated the observed centroid displacements and preserved the previously reported Euclidean displacement hierarchy:
\[
d_{\mathrm{OC3}} < d_{\mathrm{ONL}} < d_{\mathrm{OC2.5}}.
\]

Within the extended six-probe analysis, the model also approximated condition-level displacements and preserved the broad structure of the exploratory ordering. Held-out M2 and leave-condition-out analyses were used as internal tests within the same single-subject dataset. A complementary within-session analysis compared the relative positions of the six probes with respect to the ONL centroid at M1 and M2.

This work remains exploratory, retrospective, representation-dependent, and non-causal. It does not establish clinical predictive validity, causal occlusal effects, validated viability thresholds, therapeutic superiority, or generalizable patient-level prediction. Its contribution is methodological: it examines whether observed longitudinal centroid displacement can be internally approximated within a simplified supervised model.

\end{abstract}

\vspace{0.5em}

\section{Introduction}

Understanding adaptive biomechanical systems requires distinguishing between observable performance, exploratory multivariate representation, longitudinal displacement, and the possibility of approximating observed representational change. In gait analysis, spatiotemporal variables, asymmetry measures, pressure-related descriptors, and center-of-pressure behavior provide useful observable information, but they do not uniquely identify a condition-specific physiological organization.

This study represents the third empirical step in a multi-level framework for the analysis of gait dynamics under occlusal constraint. The revised Level~3 study showed that neither an aggregated scalar score nor a static exploratory UMAP embedding uniquely resolved the occlusal observational probes \cite{raynal2026level3}. The Level~4 study then examined whether condition-level representations that remained ambiguous at one session exhibited different M1--M2 centroid displacements within a common PCA coordinate system \cite{raynal2026level4}. The present article introduces Level~5, focused on internal approximation of these observed longitudinal displacements.

The preceding Level~4 analysis reported the following Euclidean centroid-displacement hierarchy in the selected PC1--PC2 projection:
\[
d_{\mathrm{OC3}} < d_{\mathrm{ONL}} < d_{\mathrm{OC2.5}}.
\]
This hierarchy was interpreted as a representation-dependent descriptive result, not as a direct physiological stability measure, a causal occlusal effect, or a therapeutic ranking.

Level~4 remained retrospective: the M1--M2 transformation was evaluated only after both sessions had been observed. It described condition-level representational change but did not determine whether that change could be approximated by a learned model.

The present work introduces a Level~5 extension. Its objective is not to predict gait performance, clinical outcome, therapeutic response, or an optimal occlusal configuration. Rather, it asks whether the observed M1--M2 transformation in the selected PCA representation can be internally approximated within the available single-subject dataset.

The approach should therefore be understood as internal predictive approximation rather than clinical prediction. The model is trained and evaluated within the same longitudinal dataset. Its purpose is to approximate observed M2 coordinates from M1 representations and to assess whether condition-level centroid displacement and its relative ordering are preserved.

A key conceptual distinction is maintained between longitudinal system evolution and observational constraint. Between M1 and M2, the participant evolved under natural daily conditions and within the context of a structured sensorimotor intervention. Occlusal conditions were applied only during gait acquisition. They are therefore treated as discrete observational probes, not as continuous causal inputs driving the eleven-week transformation.

The present article includes three complementary analyses. First, a core Level~5 analysis focuses on ONL, OC2.5, and OC3 to preserve direct continuity with Level~4. It tests whether the model reproduces the observed Euclidean displacement hierarchy:
\[
d_{\mathrm{OC3}} < d_{\mathrm{ONL}} < d_{\mathrm{OC2.5}}.
\]
Second, an extended six-probe analysis includes ONL, OBL, OSL, OC2.5, OC3, and OC3P. Its purpose is to examine whether the approximation framework can accommodate more heterogeneous observational constraints. It is not intended to define a clinical ranking.

Third, a within-session descriptive analysis compares the relative positions of the six probe centroids with respect to ONL at M1 and M2. This analysis describes relative geometry within the selected PCA representation. It does not identify physiological states and does not replace the longitudinal displacement analysis.

\begin{figure}[H]
\centering
\fbox{%
\begin{minipage}{0.93\textwidth}
\textbf{Conceptual backbone of the Level~5 framework}

\vspace{0.5em}
\textbf{1. Longitudinal predictive approximation}
\[
M1(\lambda) \longrightarrow \widehat{M2}(\lambda)
\]
compared with the observed
\[
M2(\lambda).
\]
For each observational probe $\lambda$, the model approximates the observed M1--M2 transformation in the PC1--PC2 representation. The core Level~4-aligned hierarchy is
\[
d_{\mathrm{OC3}} < d_{\mathrm{ONL}} < d_{\mathrm{OC2.5}}.
\]

\textbf{2. Within-session relative positioning of the occlusal probes}

Within each session, the six probe centroids are compared with the ONL centroid. This describes their relative positions within the selected PCA representation at M1 and M2. It does not identify physiological states and does not define a therapeutic ranking.

Thus, Level~5 does not only approximate the M1--M2 transition. It also documents how the observational probes are positioned relative to one another within the selected PCA representation at M1 and M2. Both analyses remain exploratory, internal, single-subject, and non-causal.
\end{minipage}}
\caption{Conceptual backbone of the Level~5 analysis. The longitudinal component approximates the observed M1--M2 transformation, whereas the within-session component describes relative centroid positions in the selected PCA representation.}
\label{fig:level5_backbone}
\end{figure}

To formalize this perspective, we introduce a simplified supervised model inspired by representation-learning approaches \cite{bengio2013representation,lecun2022path}. The model aims to approximate the M1--M2 coordinate transformation and to assess whether the direction and relative magnitude of centroid displacement across observational probes can be preserved.

This work is exploratory and proof-of-concept. It does not establish causal relationships, validated clinical prediction, or therapeutic decision rules. Its contribution is to test whether an observed representation-dependent longitudinal transformation can be internally approximated before asking, at a future Level~6, whether such approximation can generalize across multiple walkers.

\section{Model}

\subsection{Common PCA representation}

Let $x \in \mathbb{R}^{n}$ denote the observable biomechanical variables describing gait, and let
\[
r=\Phi(x), \qquad r\in\mathbb{R}^{m},
\]
denote a lower-dimensional representation obtained through dimensionality reduction.

In the present study, $\Phi$ was defined using Principal Component Analysis (PCA) \cite{jolliffe2016}. PCA was selected because it provides a reproducible linear coordinate system in which centroid positions and Euclidean displacements can be computed consistently.

The PCA space should not be interpreted as a direct physiological state space or as an intrinsic latent manifold. It is a model-dependent representation summarizing part of the variance contained in the observable data.

In the core Level~5 analysis, the representation was restricted to the PC1--PC2 plane to preserve computational continuity with Level~4. In the extended six-probe analysis, PCA was recomputed on the expanded dataset. Distances from the two analyses therefore belong to different coordinate systems and should not be numerically compared across analyses.

\subsection{Longitudinal transformation}

The observed transformation between two sessions is represented in the selected PCA coordinate system as
\[
r_{t+\Delta}=G_{\theta}(r_t,u),
\]
where $r_t$ denotes the representation at M1, $r_{t+\Delta}$ denotes the representation at M2, $\Delta$ is the eleven-week interval, and $u$ represents the combined longitudinal context.

The longitudinal context includes intrinsic system evolution, structured sensorimotor intervention, spontaneous variability, daily-life conditions, and other factors that cannot be separated in the present design. Consequently, $G_{\theta}$ does not represent a causal treatment-response model.

The use of $G_{\theta}$ does not imply that the true physiological dynamics of the participant have been identified. It denotes only a learned approximation of the observed coordinate transformation within the selected representation.

\subsection{Observational constraints}

Occlusal conditions were applied only during gait acquisition and were not continuously imposed between M1 and M2. They are therefore represented here as observational constraints rather than as continuous causal drivers of longitudinal evolution.

Conceptually, the expression of the system under an occlusal condition \(\lambda\) can be written as:
\[
z^{(\lambda)} = H_\phi(z,\lambda),
\]
where \(\lambda\) denotes the occlusal observational probe and \(H_\phi\) represents the expression of the latent state under that measurement constraint.

This formulation does not imply that occlusal conditions determine the longitudinal transformation of the system. Rather, they are treated as probes that reveal how the system is expressed under different measurement constraints at each time point.

This distinction is essential for the interpretation of the present study. The occlusal probes are not modeled as therapeutic interventions. They are not interpreted as continuous inputs acting throughout the eleven-week interval. Their role is limited to revealing the organization of the system during gait acquisition at M1 and M2.

\subsection{Predictive formulation}

Combining longitudinal transformation and observational constraint, the conceptual predictive formulation can be written as:
\[
\hat{z}_{t+\Delta}^{(\lambda)}
=
H_\phi\big(G_\theta(z_t,u),\lambda\big).
\]

This expression formalizes the idea that the system first undergoes a longitudinal transformation between M1 and M2, and that the transformed system is then observed under a given occlusal constraint.

In the present implementation, however, \(G_\theta\) and \(H_\phi\) were not estimated as two separately identifiable modules. Instead, their combined effect was approximated by a single simplified neural network receiving the latent state, the occlusal-condition descriptors, and the longitudinal-transition indicator as inputs.

The separation between \(G_\theta\) and \(H_\phi\) should therefore be understood as a conceptual decomposition rather than an empirically identifiable factorization in the present dataset.

Accordingly, the model does not isolate the independent effects of occlusion, intervention, time, spontaneous evolution, or neurological state. It approximates the observed M1--M2 latent transformation within the available single-subject dataset.

\subsection{Internal predictive approximation}

The term predictive is used here in a restricted methodological sense. It refers to the internal approximation of observed M1--M2 coordinate transformations, not to prospective clinical prediction.

The model receives an M1 latent coordinate and a description of the occlusal observational probe, and returns an estimated M2 latent coordinate:
\[
(z_{M1}^{(\lambda)}, c_\lambda, u)
\longmapsto
\hat{z}_{M2}^{(\lambda)},
\]
where \(z_{M1}^{(\lambda)} \in \mathbb{R}^2\) denotes the PC1--PC2 coordinate at M1, \(c_\lambda\) denotes the encoded occlusal observational probe, and \(u\) denotes the M1 \(\rightarrow\) M2 transition.

The objective is not to predict future clinical outcome. It is to test whether the observed viability-related transformation can be approximated internally and whether the displacement hierarchy identified retrospectively at Level 4 can be preserved by the model.

\subsection{Learning objective}

The model was trained to minimize the discrepancy between the observed M2 latent state and the predicted M2 latent state. For each computational M1--M2 pair, the learning objective was defined as:
\[
\mathcal{L}
=
\left\|
z_{M2}^{(\lambda)}
-
\hat{z}_{M2}^{(\lambda)}
\right\|^2,
\]
where \(z_{M2}^{(\lambda)}\) denotes the observed latent coordinate at M2 under occlusal observational probe \(\lambda\), and \(\hat{z}_{M2}^{(\lambda)}\) denotes the corresponding predicted latent coordinate.

The objective was not to reconstruct the original observable biomechanical variables, nor to predict clinical outcome. The objective was to approximate the observed M1--M2 transformation directly in representation space and to evaluate whether the relative displacement hierarchy across occlusal observational probes could be preserved.

For each condition \(\lambda\), the observed M1--M2 centroid displacement was defined as:
\[
d_{\mathrm{obs}}^{(\lambda)}
=
\left\|
\bar{z}_{M2}^{(\lambda)}
-
\bar{z}_{M1}^{(\lambda)}
\right\|.
\]

The predicted M1--M2 centroid displacement was defined as:
\[
d_{\mathrm{pred}}^{(\lambda)}
=
\left\|
\hat{\bar{z}}_{M2}^{(\lambda)}
-
\bar{z}_{M1}^{(\lambda)}
\right\|,
\]
where \(\hat{\bar{z}}_{M2}^{(\lambda)}\) denotes the centroid of the predicted M2 latent coordinates for condition \(\lambda\).

The centroid approximation error was defined as:
\[
e_{\mathrm{centroid}}^{(\lambda)}
=
\left\|
\hat{\bar{z}}_{M2}^{(\lambda)}
-
\bar{z}_{M2}^{(\lambda)}
\right\|.
\]

These quantities allowed the model to be evaluated at the condition level rather than only at the individual observation level. This distinction is important because individual M1 and M2 observations were not physiologically matched stride-by-stride. Therefore, pointwise prediction error was considered secondary, whereas centroid-level displacement and hierarchy preservation were treated as the primary interpretative outputs.

\subsection{Core, extended, and within-session analyses}

Three complementary analyses were performed.

First, a core Level 5 analysis was conducted using ONL, OC2.5, and OC3. This analysis was designed to remain directly aligned with the preceding Level 4 framework, in which longitudinal viability was assessed retrospectively through M1--M2 centroid displacement in the PC1--PC2 projection. The objective of the core analysis was to determine whether the Level 4 displacement hierarchy could be approximated by the simplified predictive latent model:
\[
d_{\mathrm{OC3}} < d_{\mathrm{ONL}} < d_{\mathrm{OC2.5}}.
\]

Second, an extended six-probe analysis was conducted using ONL, OBL, OSL, OC2.5, OC3, and OC3P. This extended analysis was not intended to redefine the Level 4 clinical hierarchy. Rather, it was used to test whether the same predictive approximation framework could accommodate additional occlusal observational probes involving open-mouth disengagement, strong voluntary clenching, and mandibular protrusion.

Third, a within-session hierarchy analysis compared the six occlusal observational probes at M1 and M2 separately by computing their distance to the ONL centroid within each session. This analysis described the relative instantaneous organization of the probes inside each session. It did not replace the longitudinal displacement analysis and was not interpreted as a therapeutic ranking.

The core analysis served as the direct continuity test between Level 4 and Level 5. The extended six-probe analysis served as an exploratory robustness analysis under more heterogeneous constraints. The within-session analysis served as a complementary descriptive analysis of within-session relative positioning at M1 and M2.

\subsection{Internal evaluation strategy}

Because the present study is based on a single participant, the model cannot establish inter-individual predictive validity. The predictive procedure was therefore interpreted as an internal approximation strategy within the available single-subject dataset.

Three levels of internal evaluation were defined.

First, the full-dataset approximation evaluated whether the model could approximate the observed M1--M2 centroid displacement when the full set of available conditions was included in the predictive modeling pipeline. This analysis tested whether the observed transformation could be represented by the model when the full condition structure was available.

Second, the held-out M2 approximation evaluated whether the model could approximate M2 observations that were withheld from training. This analysis tested internal approximation of unseen observations within the same subject and within the same occlusal condition structure. It does not test generalization to new subjects or future clinical measurements.

Third, the leave-condition-out approximation evaluated whether the model could approximate the M1--M2 transformation of an occlusal observational probe excluded from training. In this analysis, the model was trained on five occlusal conditions and evaluated on the sixth. This procedure does not test generalization to unseen patients, but it provides a more demanding internal test of whether the model can approximate the transformation of an unseen constraint within the same single-subject dataset.

For the core Level 4-aligned analysis, the main hierarchy of interest was:
\[
d_{\mathrm{OC3}} < d_{\mathrm{ONL}} < d_{\mathrm{OC2.5}}.
\]

For the extended six-probe analysis, hierarchy preservation was interpreted only as an exploratory indicator of internal consistency. Because OBL, OSL, and OC3P introduce heterogeneous constraints that are not simple VDO variants, the six-probe ordering was not interpreted as a clinical ranking of occlusal configurations.

\subsection{Relation to longitudinal displacement}

Within the present framework, Level~5 concerns approximation of an observed longitudinal descriptor rather than prediction of clinical viability. At Level~4, longitudinal behavior was summarized retrospectively through centroid displacement in a selected PCA representation. At Level~5, the same representation-dependent displacement becomes the target of a supervised approximation.

For each condition $\lambda$, the observed M1--M2 centroid displacement is
\[
d^{(\lambda)}_{\mathrm{obs}}
=
\left\|
\bar r^{(\lambda)}_{M2}
-
\bar r^{(\lambda)}_{M1}
\right\|_2.
\]
The predicted displacement is
\[
d^{(\lambda)}_{\mathrm{pred}}
=
\left\|
\widehat{\bar r}^{(\lambda)}_{M2}
-
\bar r^{(\lambda)}_{M1}
\right\|_2.
\]

These quantities are representation-dependent. They do not estimate a physiological viability region and do not constitute clinical biomarkers. Their role is to test whether the model preserves the observed condition-level displacement pattern within the selected coordinate system.

\section{Methods}

\subsection{Study design}

The data analyzed in this work come from an exploratory single-subject longitudinal design. A participant diagnosed with Parkinson's disease was assessed during two gait analysis sessions separated by eleven weeks. The first session is denoted M1 and the second session M2.

During each session, gait was recorded under six occlusal observational probes: ONL, OBL, OSL, OC2.5, OC3, and OC3P. These probes were applied only during gait acquisition and were not imposed continuously between sessions.

Between M1 and M2, the participant remained under natural daily conditions and underwent a structured sensorimotor intervention. The M1--M2 interval is therefore interpreted as a global longitudinal transformation context, including intrinsic system evolution, sensorimotor intervention, spontaneous variability, and other contextual factors that cannot be separated in the present design.

The purpose of the analysis was not to establish a causal therapeutic effect, validate an occlusal intervention, identify an optimal vertical dimension of occlusion, or predict clinical outcome. Rather, the objective was to test whether the observed latent transformation between M1 and M2 could be internally approximated within a simplified predictive representation-space framework.

\subsection{Participant and clinical context}

The participant was a Parkinsonian walker with autonomous gait. The neurological condition provides a relevant context for studying adaptive gait dynamics, sensorimotor regulation, and longitudinal reorganization, but it is not used here as an explanatory variable.

The study should be understood as a proof-of-concept single-case analysis. The results cannot be generalized to a Parkinsonian population and do not establish patient-level predictive validity.

\subsection{Sophrology-oriented sensorimotor intervention}

Between M1 and M2, the participant underwent eleven sessions of sophrology-oriented sensorimotor intervention delivered by a certified sophrologist and dental assistant. In the present Level 5 framework, this intervention is considered part of the structured longitudinal context in which the M1--M2 transformation occurred, rather than an isolated causal factor.

The intervention contributed to the clinical context of the observed transformation through body awareness, interoceptive attention, postural regulation, breathing control, and proprioceptive engagement. Its role in the present analysis is not to explain the observed representational displacement directly, but to define part of the structured transformation context between M1 and M2.

Because the intervention was not controlled against a comparison condition, its specific contribution cannot be isolated from spontaneous evolution, occlusal observational probe, neurological variability, or their possible interactions.

\subsection{Occlusal observational probes}

The complete experimental dataset included six occlusal observational probes recorded at both M1 and M2:

\begin{itemize}
\item ONL: neutral natural occlusion;
\item OBL: wide open-mouth condition without dental contact;
\item OSL: strong voluntary dental clenching;
\item OC2.5: 2.5-degree increase in vertical dimension of occlusion in centric relation;
\item OC3: 3-degree increase in vertical dimension of occlusion in centric relation;
\item OC3P: 3-degree increase in vertical dimension of occlusion combined with mandibular protrusion, corresponding to a 2 mm displacement of the temporomandibular hinge-axis position.
\end{itemize}

The primary interpretative reference remained the Level 4 core set composed of ONL, OC2.5, and OC3. These three conditions were selected because they represent clinically comparable VDO-related constraints in centric relation and allow direct continuity with the preceding Level 4 analysis.

The additional conditions OBL, OSL, and OC3P were not interpreted as simple VDO variants. OBL introduces open-mouth occlusal disengagement, OSL introduces strong voluntary muscular contraction, and OC3P combines vertical increase with mandibular protrusion and hinge-axis displacement. These conditions were therefore analyzed as extended observational probes, useful for testing whether the predictive approximation framework could accommodate more heterogeneous constraints.

All six occlusal conditions were treated as observational probes applied during measurement, not as continuous causal drivers of longitudinal system evolution between M1 and M2.

\subsection{Data acquisition}

Gait data were acquired using instrumented insoles providing spatiotemporal parameters and pressure-related variables. Each recording produced multivariate observations describing gait behavior under a given occlusal observational probe and session.

The dataset consisted of twelve recordings corresponding to six occlusal observational probes across two measurement sessions:
\[
\text{M1-ONL}, \text{M1-OBL}, \text{M1-OSL}, \text{M1-OC2.5}, \text{M1-OC3}, \text{M1-OC3P},
\]
\[
\text{M2-ONL}, \text{M2-OBL}, \text{M2-OSL}, \text{M2-OC2.5}, \text{M2-OC3}, \text{M2-OC3P}.
\]

Although the file metadata may contain session dates, the clinical interval between M1 and M2 is considered here as eleven weeks for all conditions.

Only the numerical biomechanical variables retained after preprocessing were used for representation-space modeling.

\subsection{Dataset structure}

The number of observations differed across occlusal observational probes and sessions. The available observations were:

\begin{table}[H]
\centering
\caption{
Dataset structure across the six occlusal observational probes and the two measurement sessions.
The M1--M2 interval was interpreted as an eleven-week longitudinal transformation for all conditions.
}
\label{tab:dataset-structure}
\begin{tabular}{lcc}
\hline
Condition & M1 observations & M2 observations \\
\hline
ONL   & 50 & 60 \\
OBL   & 33 & 57 \\
OSL   & 46 & 58 \\
OC2.5 & 41 & 60 \\
OC3   & 51 & 57 \\
OC3P  & 49 & 62 \\
\hline
\end{tabular}
\end{table}

\subsection{Preprocessing}

Variables were harmonized across all recordings. Categorical fields and non-informative technical fields were excluded from the numerical matrix. For direct consistency with the preceding Level 4 analysis, the core PCA projection was recomputed using the same numerical feature set retained in that pipeline, including temporal numerical columns when present. The excluded fields were limited to non-numerical or non-informative technical variables, including side, flag, and overflow.

The retained numerical matrix contained 60 variables. These variables were standardized using z-score normalization across the aggregated dataset before PCA projection.

For the core Level 5 analysis, preprocessing was kept directly aligned with the preceding Level 4 analysis in order to preserve the interpretability of the core ONL--OC2.5--OC3 displacement hierarchy.

For the extended six-probe analysis, the PCA projection was recomputed using the complete six-probe dataset. Because PCA depends on the data matrix used to estimate the projection axes, distances obtained in the six-probe projection were interpreted as belonging to the extended analysis and were not used to replace the core Level 4-aligned distances.

\subsection{Latent-space construction}

Dimensionality reduction was performed using PCA \cite{jolliffe2016}. The PCA model was fitted on the aggregated standardized dataset corresponding to the analysis being performed.

For the core analysis, the retained representation space was the PC1--PC2 plane corresponding to the Level 4-aligned ONL, OC2.5, and OC3 analysis.

For the extended six-probe analysis, PCA was recomputed on the aggregated standardized dataset including ONL, OBL, OSL, OC2.5, OC3, and OC3P at both M1 and M2. The retained representation space was again restricted to PC1--PC2.

Each observation was therefore represented as:
\[
z_i =
\begin{pmatrix}
PC1_i \\
PC2_i
\end{pmatrix}.
\]

Centroids were then computed for each occlusal observational probe and session.

\subsection{Construction of computational M1--M2 pairs}

Training pairs were constructed by associating observations from M1 and M2 within the same occlusal observational probe. For each condition \(\lambda\), pairs of the form
\[
(z_{M1}^{(\lambda)}, z_{M2}^{(\lambda)})
\]
were generated, where \(z_{M1}^{(\lambda)}\) denotes an observation projected in representation space at M1 and \(z_{M2}^{(\lambda)}\) denotes an observation projected in representation space at M2.

Because individual observations at M1 and M2 do not constitute physiologically matched pairs, these training pairs were constructed only for computational purposes. Within each occlusal observational probe, observations were aligned after truncating to the minimum available number of observations between M1 and M2.

This pairing should not be interpreted as stride-level longitudinal correspondence. It is a pragmatic procedure for approximating condition-level M1--M2 transformations in representation space.

The time interval between M1 and M2 was treated as a single longitudinal transition step. Intermediate dynamics were not modeled.

\subsection{Encoding of occlusal observational probes}

For the extended six-probe analysis, occlusal observational probes were encoded using a structured descriptor-based representation rather than a purely categorical one-hot representation.

A purely one-hot encoding would not allow meaningful leave-condition-out testing, because a withheld condition would correspond to an unseen categorical input. For this reason, the extended six-probe analysis used descriptors intended to represent the mechanical and functional characteristics of each observational probe.

The descriptor set included:

\begin{itemize}
\item presence or absence of dental contact;
\item open-mouth disengagement;
\item strong voluntary clenching;
\item vertical dimension increase;
\item mandibular protrusion.
\end{itemize}

The descriptor-based encoding used in the predictive model is shown in Table~\ref{tab:occlusal-encoding}.

\begin{table}[H]
\centering
\caption{
Descriptor-based encoding of the occlusal observational probes used for the extended six-probe predictive approximation.
This encoding was used to allow leave-condition-out testing and should not be interpreted as a validated clinical scoring system.
}
\label{tab:occlusal-encoding}
\begin{tabular}{lccccc}
\hline
    Condition & Dental contact & Open mouth & Strong clenching & VDO increase & Protrusion mm \\
\hline
ONL   & 1 & 0 & 0 & 0.0 & 0.0 \\
OBL   & 0 & 1 & 0 & 0.0 & 0.0 \\
OSL   & 1 & 0 & 1 & 0.0 & 0.0 \\
OC2.5 & 1 & 0 & 0 & 2.5 & 0.0 \\
OC3   & 1 & 0 & 0 & 3.0 & 0.0 \\
OC3P  & 1 & 0 & 0 & 3.0 & 2.0 \\
\hline
\end{tabular}
\end{table}

This descriptor-based encoding was used only to support internal predictive approximation. It should not be interpreted as a validated biomechanical parameterization of occlusal state, nor as a clinical scoring system.

\subsection{Predictive model implementation}

A simplified supervised machine-learning model was implemented to approximate the observed M1--M2 transformation in representation space. The model consisted of a feed-forward neural network trained on computational M1--M2 pairs. Its use was restricted to internal approximation within the same single-subject dataset and should not be interpreted as a clinical prediction model.

For each computational pair, the model input consisted of the M1 latent coordinates and the encoded occlusal observational probe:
\[
(z_{M1}^{(\lambda)}, c_\lambda, u),
\]
where \(z_{M1}^{(\lambda)} \in \mathbb{R}^2\) denotes the PC1--PC2 latent coordinate at M1, \(c_\lambda\) denotes the descriptor-based representation of the occlusal observational probe, and \(u\) denotes the M1 \(\rightarrow\) M2 longitudinal transition.

The model output was the predicted M2 latent coordinate:
\[
\hat{z}_{M2}^{(\lambda)} \in \mathbb{R}^2.
\]

The machine-learning task was therefore supervised in a restricted methodological sense: the observed M2 latent coordinates served as training targets for approximating the M1--M2 latent transformation. The objective was not to infer a general predictive rule, but to test whether the observed transformation could be approximated internally in the selected representation space.

The model approximated the combined effect of longitudinal transformation and observational constraint. It did not estimate \(G_\theta\) and \(H_\phi\) as two separately identifiable modules.

\subsection{Training procedure}

The model was trained using mean squared error loss between observed and predicted M2 latent coordinates:
\[
\mathcal{L}
=
\left\|
z_{M2}^{(\lambda)}
-
\hat{z}_{M2}^{(\lambda)}
\right\|^2.
\]

Optimization was performed using the Adam optimizer.

The objective of training was not to achieve optimal predictive performance, compare neural network architectures, or demonstrate generalizable prediction. The objective was to assess whether the observed M1--M2 latent transformation could be approximated by a simple nonlinear mapping within the available single-subject dataset.

Accordingly, the term predictive is used in a restricted sense: it refers to internal approximation of observed longitudinal transformations in representation space, not to prospective clinical prediction.

\begin{table}[H]
\centering
\caption{
Reproducibility parameters of the simplified supervised machine-learning model used for internal predictive approximation.
}
\label{tab:model-reproducibility}
\begin{tabular}{ll}
\hline
Item & Specification \\
\hline
Latent space & PC1--PC2 PCA projection \\
Learning framework & Supervised machine learning \\
Model type & Feed-forward neural network \\
Input variables & PC1, PC2, occlusal-probe descriptors, transition indicator \\
Output variables & Predicted M2 PC1, predicted M2 PC2 \\
Training target & Observed M2 PC1, observed M2 PC2 \\
Loss function & Mean squared error \\
Optimizer & Adam \\
Hidden layers & 16, 16 \\
Activation function & ReLU \\
Learning rate & 0.001 \\
Epochs & 800 \\
Batch type & Full-batch gradient update \\
Random seed & 42 \\
Held-out split & 20\% within condition \\
Feature set & 60 numerical variables; side, flag, and overflow excluded \\
Temporal numerical columns & Included for Level 4-aligned consistency \\
\hline
\end{tabular}
\end{table}

\subsection{Reproducibility and computational implementation}

All analyses were implemented in Python using a reproducible computational pipeline. The script included data loading, numerical feature selection, standardization, PCA projection, centroid computation, construction of computational M1--M2 pairs, supervised neural-network training, full-dataset approximation, held-out M2 evaluation, leave-condition-out evaluation, and export of derived result tables.

The supervised machine-learning component was used only to approximate the observed M1--M2 transformation in the PC1--PC2 representation space. It was not designed to optimize clinical prediction performance, compare machine-learning architectures, or generate a patient-level forecasting model.

To facilitate reproducibility, the Python script and the anonymized derived datasets required to reproduce the analyses may be made available by the corresponding author upon reasonable request.

\subsection{Evaluation strategy}

Three levels of evaluation were used.

First, the full-dataset approximation assessed whether the model could approximate observed centroid displacement when all available computational M1--M2 pairs were included in the modeling pipeline. This analysis evaluated whether the observed transformation could be represented when the full condition structure was available to the model.

Second, the held-out M2 approximation assessed whether the model could approximate M2 observations withheld from training. Observations were split within condition into training and held-out subsets. The model was trained on the training subset and evaluated on held-out M2 targets. This analysis tested internal approximation of unseen observations within the same single-subject dataset.

Third, the leave-condition-out approximation assessed whether the model could approximate the M1--M2 transformation of one occlusal observational probe excluded from training. In this analysis, the model was trained on five occlusal observational probes and evaluated on the sixth. This procedure does not test generalization to unseen patients, but it provides a more demanding internal test of whether the model can approximate the transformation of an unseen constraint within the same participant.

For each analysis, the primary evaluation level was the centroid level. Pointwise prediction error was also computed but was treated as secondary because M1 and M2 observations were not physiologically matched stride-by-stride.

\subsection{Outcome measures}

For each occlusal observational probe \(\lambda\), three centroid-level quantities were computed.

The observed M1--M2 displacement was:
\[
d_{\mathrm{obs}}^{(\lambda)}
=
\left\|
\bar{z}_{M2}^{(\lambda)}
-
\bar{z}_{M1}^{(\lambda)}
\right\|.
\]

The predicted M1--M2 displacement was:
\[
d_{\mathrm{pred}}^{(\lambda)}
=
\left\|
\hat{\bar{z}}_{M2}^{(\lambda)}
-
\bar{z}_{M1}^{(\lambda)}
\right\|.
\]

The centroid approximation error was:
\[
e_{\mathrm{centroid}}^{(\lambda)}
=
\left\|
\hat{\bar{z}}_{M2}^{(\lambda)}
-
\bar{z}_{M2}^{(\lambda)}
\right\|.
\]

In addition to longitudinal M1--M2 displacement, a within-session hierarchy of the six occlusal observational probes was computed at M1 and M2. For each session, the ONL centroid was used as the intra-session reference. The distance of each probe centroid to the ONL centroid was defined as:
\[
d_{M1}^{(\lambda|ONL)}
=
\left\|
\bar{z}_{M1}^{(\lambda)}
-
\bar{z}_{M1}^{(ONL)}
\right\|,
\]
and
\[
d_{M2}^{(\lambda|ONL)}
=
\left\|
\bar{z}_{M2}^{(\lambda)}
-
\bar{z}_{M2}^{(ONL)}
\right\|.
\]

This analysis was used to describe the relative organization of the six occlusal observational probes within each measurement session. It should not be interpreted as a therapeutic ranking. ONL was used as a reference condition, not as a validated optimum.

Pointwise prediction error was summarized using the root mean squared error when appropriate.

For the core Level 4-aligned analysis, the main hierarchy of interest was:
\[
d_{\mathrm{OC3}} < d_{\mathrm{ONL}} < d_{\mathrm{OC2.5}}.
\]

For the extended six-probe analysis, preservation of the global ordering was interpreted as an exploratory indicator of internal consistency, not as a clinical ranking.

\subsection{Interpretation framework}

All results were interpreted within an exploratory, retrospective, and non-causal framework. The model was not intended to predict future clinical outcome, identify an optimal occlusal configuration, or establish a causal relationship between vertical dimension of occlusion and gait dynamics.

Occlusal observational probes were interpreted as constraints applied during measurement. The M1--M2 transformation was interpreted as the combined outcome of intrinsic system evolution, structured sensorimotor intervention, spontaneous variability, and contextual factors.

The Level 5 analysis therefore extends Level 4 by testing whether observed viability-related representational displacement can be internally approximated within a simplified predictive representation-space framework.

\subsection{Methodological scope}

This analysis is limited to a single participant and a single M1--M2 transition. The model was evaluated within the same available dataset used to construct the latent transformation. Therefore, the results should be interpreted as proof-of-concept evidence that the observed transformation is approximable in representation space, not as evidence of generalizable predictive validity.

The method does not provide an operational clinical decision rule, does not validate a biomarker of viability, and does not establish a therapeutic optimum.

\section{Results}

\subsection{Core Level 5 analysis aligned with the preceding Level 4 framework}

The core Level 5 analysis focused on ONL, OC2.5, and OC3 in order to preserve direct continuity with the preceding Level 4 analysis.

The preceding Level 4 analysis reported the displacement hierarchy:
\[
d_{\mathrm{OC3}} < d_{\mathrm{ONL}} < d_{\mathrm{OC2.5}},
\]
with centroid displacements close to 5.76 for ONL, 6.47 for OC2.5, and 5.32 for OC3 in the Level 4 projection.

Using the Level 5 predictive pipeline with the Level 4-aligned numerical feature set, the recomputed observed displacements were:
\[
d_{\mathrm{ONL}} = 5.73,
\qquad
d_{\mathrm{OC2.5}} = 6.39,
\qquad
d_{\mathrm{OC3}} = 5.35.
\]

The ordering was unchanged:
\[
d_{\mathrm{OC3}} < d_{\mathrm{ONL}} < d_{\mathrm{OC2.5}}.
\]

Thus, OC3 showed the lowest longitudinal centroid displacement, ONL occupied an intermediate position, and OC2.5 showed the highest displacement. This reproduces the hierarchy reported in the preceding Level 4 analysis and provides the observational reference for the present Level 5 approximation.

These values should not be interpreted as direct clinical effects of occlusal configuration. They describe centroid displacement in the selected PCA projection under the combined longitudinal context of occlusal probing, sensorimotor intervention, spontaneous system evolution, and single-subject variability.

\subsection{Full-dataset predictive approximation of the core conditions}

The simplified predictive model was first evaluated on the core Level 4-aligned conditions. The objective was to assess whether the model could approximate the observed M1--M2 centroid displacement and preserve the Level 4 hierarchy.

Predicted centroid displacements were:
\[
\hat{d}_{\mathrm{ONL}} = 5.72,
\qquad
\hat{d}_{\mathrm{OC2.5}} = 6.45,
\qquad
\hat{d}_{\mathrm{OC3}} = 5.32.
\]

The predicted ordering was:
\[
\hat{d}_{\mathrm{OC3}} < \hat{d}_{\mathrm{ONL}} < \hat{d}_{\mathrm{OC2.5}}.
\]

The model therefore preserved the relative hierarchy of longitudinal centroid displacement observed in the Level 4 analysis. OC3 remained associated with the lowest predicted displacement, ONL with an intermediate predicted displacement, and OC2.5 with the highest predicted displacement.

This agreement should be interpreted as a proof-of-concept approximation of the observed latent transformation, not as evidence that the model can predict future clinical evolution in unseen data.

\begin{table}[H]
\centering
\caption{
Observed and predicted centroid displacements in the core Level 5 approximation.
The observed displacements were recomputed in the Level 5 predictive pipeline using the Level 4-aligned numerical feature set.
The displacement hierarchy remains consistent with the preceding Level 4 analysis:
\(d_{\mathrm{OC3}} < d_{\mathrm{ONL}} < d_{\mathrm{OC2.5}}\).
Centroid error corresponds to the distance between observed and predicted M2 centroids.
}
\label{tab:core-displacements}
\begin{tabular}{lccc}
\hline
Condition & Observed displacement & Predicted displacement & Centroid error \\
\hline
ONL   & 5.73 & 5.72 & 0.01 \\
OC2.5 & 6.39 & 6.45 & 0.06 \\
OC3   & 5.35 & 5.32 & 0.03 \\
\hline
\end{tabular}
\end{table}

\begin{figure}[H]
\centering
\includegraphics[width=0.85\textwidth]{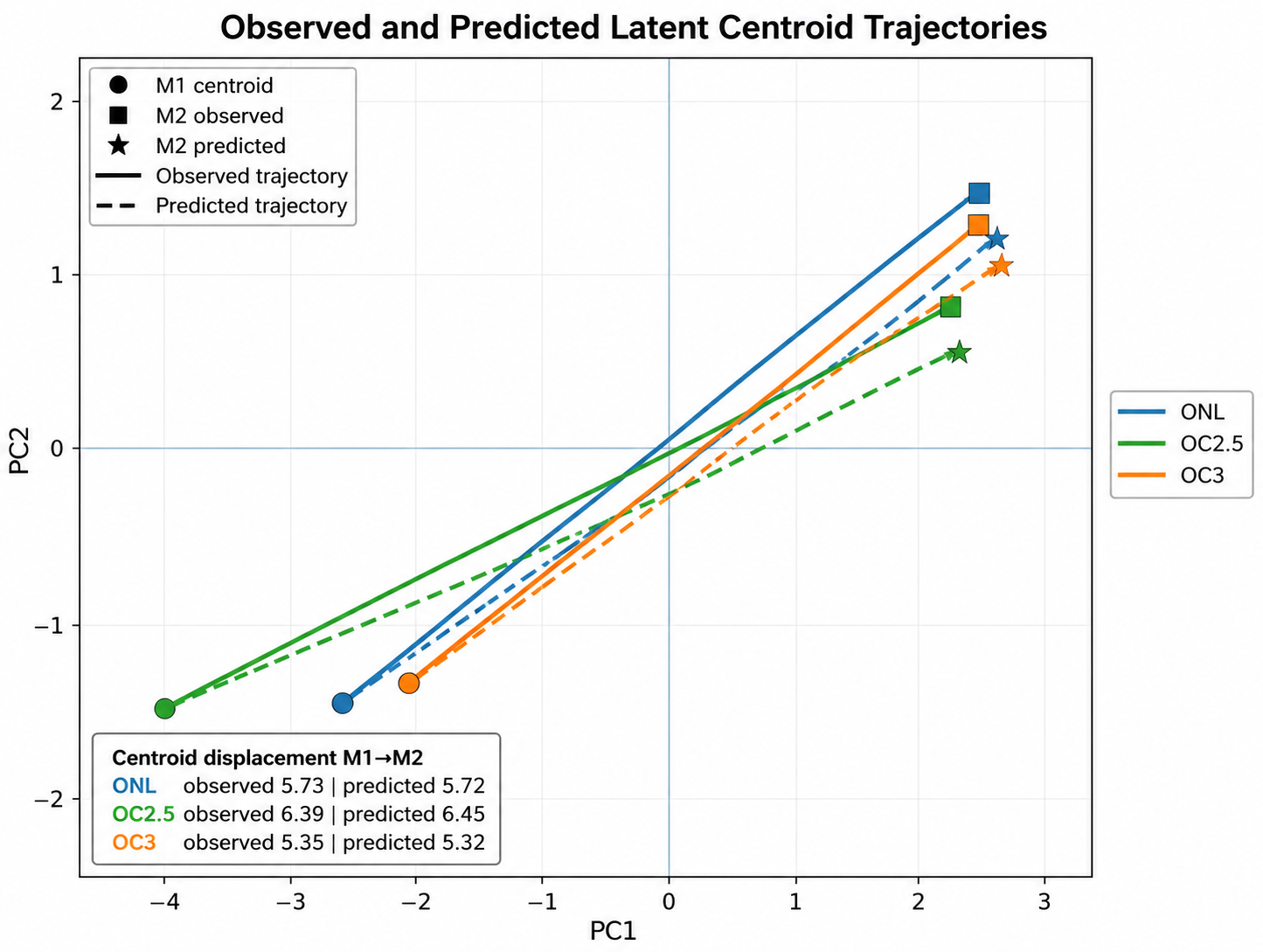}
\caption{
Observed and predicted latent centroid trajectories between M1 and M2 for the core Level 4-aligned conditions.
Solid lines represent observed M1--M2 centroid trajectories, whereas dashed lines represent model-predicted trajectories.
The preceding Level 4 analysis reported centroid displacements close to 5.76 for ONL, 6.47 for OC2.5, and 5.32 for OC3.
Using the Level 5 predictive pipeline, recomputed observed centroid displacements were 5.73 for ONL, 6.39 for OC2.5, and 5.35 for OC3.
Predicted centroid displacements were 5.72 for ONL, 6.45 for OC2.5, and 5.32 for OC3.
The displacement hierarchy \(d_{\mathrm{OC3}} < d_{\mathrm{ONL}} < d_{\mathrm{OC2.5}}\) was preserved.
These predicted trajectories are shown as a proof-of-concept approximation of the observed latent transformation, not as evidence of generalizable clinical prediction.
}
\label{fig:core-centroid-trajectories}
\end{figure}

\subsection{Extended six-probe representational displacement analysis}

The extended analysis included the six occlusal observational probes recorded at both M1 and M2: ONL, OBL, OSL, OC2.5, OC3, and OC3P.

When the PCA projection was recomputed using the six-probe dataset, the observed M1--M2 centroid displacements were:
\[
d_{\mathrm{ONL}} = 5.72,
\qquad
d_{\mathrm{OBL}} = 6.72,
\qquad
d_{\mathrm{OSL}} = 5.57,
\]
\[
d_{\mathrm{OC2.5}} = 6.32,
\qquad
d_{\mathrm{OC3}} = 5.77,
\qquad
d_{\mathrm{OC3P}} = 5.78.
\]

The corresponding exploratory ordering in the six-probe projection was:
\[
d_{\mathrm{OSL}} < d_{\mathrm{ONL}} < d_{\mathrm{OC3}} < d_{\mathrm{OC3P}} < d_{\mathrm{OC2.5}} < d_{\mathrm{OBL}}.
\]

Because OC3 and OC3P showed very close displacement values, their relative position should not be overinterpreted.

This ordering should not be interpreted as a clinical ranking of occlusal conditions. Unlike the core Level 4 set, the six-probe analysis includes heterogeneous constraints. OBL introduces open-mouth disengagement, OSL introduces strong voluntary clenching, and OC3P combines vertical increase with mandibular protrusion and hinge-axis displacement.

The purpose of the extended analysis was therefore not to redefine the Level 4 viability hierarchy, but to test whether the predictive approximation framework could be applied to a broader set of observational constraints.

\subsection{Within-session hierarchy of occlusal observational probes at M1 and M2}

In addition to the longitudinal M1--M2 displacement analysis, the six occlusal observational probes were compared within each measurement session. For this analysis, the ONL centroid was used as the intra-session reference, and the Euclidean distance between each probe centroid and the ONL centroid was computed in the same PC1--PC2 projection.

This analysis addresses a question distinct from longitudinal viability. The longitudinal displacement analysis evaluates how far each condition moved between M1 and M2. The within-session analysis evaluates how far each occlusal probe was positioned from ONL within M1 and within M2 separately.

At M1, the hierarchy of distances to ONL was:
\[
ONL < OSL < OC3 < OC3P < OC2.5 < OBL.
\]

At M2, the hierarchy was:
\[
ONL < OSL < OC3 < OBL < OC3P < OC2.5.
\]

The first three positions were preserved across the two sessions: ONL, OSL, and OC3 remained the three probes closest to the ONL reference. OC3 therefore remained close to ONL at both M1 and M2 within the extended six-probe latent projection. OC2.5 remained among the most distant probes, ranking fifth at M1 and sixth at M2. OBL showed the largest change in relative position, being the most distant probe from ONL at M1 but moving closer to ONL at M2.

These within-session hierarchies should be interpreted descriptively. They do not define therapeutic superiority and do not replace the longitudinal displacement hierarchy. Rather, they describe the relative latent organization of the six probes within each measurement session.


\begin{table}[H]
\centering
\caption{
Within-session hierarchy of the six occlusal observational probes at M1 and M2.
Distances were computed between each condition centroid and the ONL centroid within the same PC1--PC2 latent projection.
The longitudinal displacement corresponds to the distance between the M1 and M2 centroids of the same condition.
}
\label{tab:within-session-hierarchy}
\small
\begin{tabular}{lcccccc}
\hline
Condition & M1 dist. & M1 rank & M2 dist. & M2 rank & M1--M2 disp. & Long. rank \\
\hline
ONL   & 0.00 & 1 & 0.00 & 1 & 5.72 & 2 \\
OBL   & 1.74 & 6 & 0.32 & 4 & 6.72 & 6 \\
OSL   & 0.16 & 2 & 0.07 & 2 & 5.57 & 1 \\
OC2.5 & 1.10 & 5 & 0.51 & 6 & 6.32 & 5 \\
OC3   & 0.26 & 3 & 0.22 & 3 & 5.77 & 3 \\
OC3P  & 0.42 & 4 & 0.33 & 5 & 5.78 & 4 \\
\hline
\end{tabular}
\normalsize
\end{table}

\subsection{Predictive approximation across the six observational probes}

In the full six-probe analysis, the predictive model approximated the observed M1--M2 centroid displacements across all six occlusal observational probes.

Predicted centroid displacements were:
\[
\hat{d}_{\mathrm{ONL}} = 5.61,
\qquad
\hat{d}_{\mathrm{OBL}} = 6.65,
\qquad
\hat{d}_{\mathrm{OSL}} = 5.55,
\]
\[
\hat{d}_{\mathrm{OC2.5}} = 6.41,
\qquad
\hat{d}_{\mathrm{OC3}} = 5.75,
\qquad
\hat{d}_{\mathrm{OC3P}} = 5.74.
\]

The predicted ordering was:
\[
\hat{d}_{\mathrm{OSL}} < \hat{d}_{\mathrm{ONL}} < \hat{d}_{\mathrm{OC3P}} < \hat{d}_{\mathrm{OC3}} < \hat{d}_{\mathrm{OC2.5}} < \hat{d}_{\mathrm{OBL}}.
\]

Thus, within the full-dataset six-probe approximation, the model preserved the global structure of the exploratory ordering. The lowest displacement remained associated with OSL, and the highest displacements remained associated with OC2.5 and OBL. OC3 and OC3P remained closely grouped, with a minor ordering inversion between these two nearly equivalent probes. This inversion should not be overinterpreted because their observed displacements were very close.

Centroid errors remained low across the six probes, ranging from 0.01 to 0.20. The smallest centroid error was observed for OSL, and the largest for OC2.5. These values indicate close centroid-level approximation within the available dataset, but do not establish generalizable predictive validity.

\begin{table}[H]
\centering
\caption{
Observed and predicted centroid displacements across the six occlusal observational probes in the extended PCA projection.
The model preserved the global structure of the six-probe displacement pattern, although OC3 and OC3P showed a minor ordering inversion despite nearly equivalent observed and predicted displacements.
The core Level 4 hierarchy remains interpreted only for ONL, OC2.5, and OC3.
OBL, OSL, and OC3P were included as extended observational probes and were not interpreted as simple VDO variants.
}
\label{tab:six-probe-displacements}
\begin{tabular}{lccc}
\hline
Condition & Observed displacement & Predicted displacement & Centroid error \\
\hline
ONL   & 5.72 & 5.61 & 0.07 \\
OBL   & 6.72 & 6.65 & 0.04 \\
OSL   & 5.57 & 5.55 & 0.01 \\
OC2.5 & 6.32 & 6.41 & 0.20 \\
OC3   & 5.77 & 5.75 & 0.12 \\
OC3P  & 5.78 & 5.74 & 0.03 \\
\hline
\end{tabular}
\end{table}

In the observed six-probe displacement hierarchy, OC3 was slightly lower than OC3P. In the predicted hierarchy, OC3P was slightly lower than OC3. This minor inversion should not be overinterpreted because the two displacement values were nearly identical in both observed and predicted analyses.

\subsection{Held-out M2 approximation}

To move beyond full-dataset approximation, a held-out M2 analysis was performed. In this analysis, part of the M2 observations was withheld from training within each condition, and the model was evaluated on these held-out M2 targets.

The global held-out pointwise root mean squared error was:
\[
\mathrm{RMSE}_{\mathrm{held\text{-}out}} = 1.78.
\]

At the condition level, centroid errors ranged from 0.04 to 1.70. The smallest centroid error was observed for OBL, whereas the largest was observed for OC3P. Pointwise RMSE values were higher for ONL, OBL, and OC3P, consistent with the interpretation that some probes introduce more heterogeneous constraint-dependent responses.

\begin{table}[H]
\centering
\caption{
Held-out M2 approximation.
The model was trained on a subset of the available observations and evaluated on M2 observations withheld from training within the same single-subject dataset.
Observed and predicted displacements were computed at the held-out centroid level.
This analysis does not test generalization to unseen subjects.
}
\label{tab:held-out-m2}
\begin{tabular}{lcccc}
\hline
Condition & Held-out observations & Observed displacement & Predicted displacement & Centroid error \\
\hline
ONL   & 10 & 6.31 & 6.28 & 0.78 \\
OBL   & 7  & 6.20 & 6.16 & 0.04 \\
OSL   & 10 & 6.12 & 6.18 & 1.18 \\
OC2.5 & 9  & 6.73 & 6.58 & 0.65 \\
OC3   & 11 & 5.78 & 5.70 & 0.21 \\
OC3P  & 10 & 5.73 & 5.77 & 1.70 \\
\hline
\end{tabular}
\end{table}

\begin{table}[H]
\centering
\caption{
Pointwise RMSE for held-out M2 observations.
Pointwise errors were interpreted as secondary outcomes because M1 and M2 observations were not physiologically matched stride-by-stride.
}
\label{tab:held-out-rmse}
\begin{tabular}{lc}
\hline
Condition & Pointwise RMSE \\
\hline
ONL   & 2.15 \\
OBL   & 2.53 \\
OSL   & 1.25 \\
OC2.5 & 1.84 \\
OC3   & 0.50 \\
OC3P  & 2.04 \\
\hline
\end{tabular}
\end{table}

\subsection{Leave-condition-out internal approximation}

A leave-condition-out analysis was then performed to test whether the model could approximate the M1--M2 transformation of an occlusal observational probe excluded from training. For each row of Table~\ref{tab:leave-condition-out}, the model was trained on five occlusal observational probes and evaluated on the withheld sixth probe.

The leave-condition-out approximation was more demanding than the full-dataset and held-out M2 analyses. Centroid errors ranged from 0.40 to 1.08. The lowest centroid error was observed for OC3P, followed by OC3 and OBL. Higher centroid errors were observed for ONL, OC2.5, and OSL.

These results indicate that leave-condition-out approximation remained possible within the single-subject dataset, but was less stable than full-dataset approximation. The errors should not be interpreted as failures of clinical prediction, because the purpose of this analysis was limited to internal approximation of an unseen observational probe within the same participant.

\begin{table}[H]
\centering
\caption{
Leave-condition-out internal approximation.
For each row, the model was trained on five occlusal observational probes and evaluated on the withheld probe.
This analysis tests internal approximation of an unseen condition within the same single-subject dataset, not generalization to unseen patients.
}
\label{tab:leave-condition-out}
\begin{tabular}{lcccc}
\hline
Held-out condition & Observed displacement & Predicted displacement & Centroid error & Pointwise RMSE \\
\hline
ONL   & 5.68 & 4.99 & 1.08 & 3.18 \\
OBL   & 6.66 & 6.60 & 0.58 & 2.55 \\
OSL   & 5.54 & 5.55 & 0.73 & 3.86 \\
OC2.5 & 6.27 & 6.97 & 0.78 & 2.05 \\
OC3   & 5.80 & 5.30 & 0.51 & 0.86 \\
OC3P  & 5.76 & 6.04 & 0.40 & 3.17 \\
\hline
\end{tabular}
\end{table}

\subsection{Pointwise prediction error}

At the level of individual observations, prediction error remained present. This indicates that the model did not reproduce the full fine-grained variability of the latent observations.

This result is expected given the single-subject dataset, the simplified model architecture, and the fact that M1 and M2 observations do not represent physiologically matched stride-level pairs.

The pointwise prediction error should therefore not be interpreted as a validated measure of clinical prediction accuracy. In the present study, the relevant level of interpretation is the centroid-level approximation of condition-wise representational displacement and the preservation of displacement hierarchies.

\subsection{Summary of results}

Taken together, the results show that the observed M1--M2 coordinate transformation can be internally approximated within the available single-subject dataset by a simplified supervised model.

In the core Level~4-aligned analysis, the model preserved the Euclidean centroid-displacement hierarchy reported in the preceding study:
\[
d_{\mathrm{OC3}} < d_{\mathrm{ONL}} < d_{\mathrm{OC2.5}}.
\]
The slight numerical differences between the Level~4 values and the recomputed Level~5 values reflect recalculation within the predictive pipeline. They do not alter the ordering but confirm that the result is specific to the selected preprocessing and PCA representation.

In the extended six-probe analysis, the model preserved the broad structure of the exploratory displacement ordering. Because this PCA projection was recomputed using six heterogeneous probes, the resulting numerical distances do not replace the core Level~4 analysis and should not be interpreted as a clinical hierarchy.

The within-session analysis showed differences in relative centroid position with respect to ONL. These findings describe geometry within the selected representation and do not establish proximity within the selected PCA representation or distinct physiological organization.

The results support only the restricted conclusion that observed representation-dependent longitudinal transformations can be internally approximated within the present single-subject dataset.

\section{Discussion}

The present study introduces a Level~5 extension of the multi-level framework developed in the preceding articles. The revised Level~3 study showed that neither scalar aggregation nor a static exploratory embedding uniquely resolved the observational probes \cite{raynal2026level3}. The Level~4 study then described condition-dependent M1--M2 centroid displacement within a common PCA representation \cite{raynal2026level4}. The present article asks whether that observed representation-dependent transformation can be internally approximated by a simplified supervised model.

The principal finding is that, within the available single-subject dataset, the model approximated condition-level M1--M2 transformations in the selected PCA coordinate system and preserved the core Euclidean displacement hierarchy:
\[
d_{\mathrm{OC3}} < d_{\mathrm{ONL}} < d_{\mathrm{OC2.5}}.
\]
This result does not demonstrate generalizable prediction, physiological trajectory recovery, or clinical forecasting. It shows only that the observed coordinate transformation contains a condition-level structure that can be approximated internally within the same dataset.

\subsection{From Level 4 observation to Level 5 approximation}

The preservation of the displacement hierarchy links Level~5 computationally to Level~4. At Level~4, lower Euclidean centroid displacement was used as a representation-dependent proxy for lower observed longitudinal reorganization. The covariance-normalized analysis showed that this hierarchy was not invariant to the distance metric.

Level~5 therefore does not predict physiological viability. It approximates the Euclidean displacement pattern selected operationally at Level~4.

The contribution is methodological rather than clinical. It shows that an observed, representation-dependent longitudinal summary can be encoded and approximated by a supervised model, while leaving open whether that summary corresponds to clinically meaningful viability.

\subsection{Core analysis and extended six-probe analysis}

The core analysis was intentionally restricted to ONL, OC2.5, and OC3 in order to remain directly aligned with the preceding Level 4 article. These conditions form the continuity axis of the multi-level framework. In this core analysis, the observed displacement hierarchy was preserved by the predictive approximation.

The extended six-probe analysis served a different purpose. By including OBL, OSL, and OC3P, the model was tested against a broader set of observational probes. These additional probes are not simple VDO variants. OBL introduces open-mouth disengagement, OSL introduces strong voluntary clenching, and OC3P combines vertical increase with mandibular protrusion and hinge-axis displacement.

Within this extended analysis, the model preserved the global structure of the exploratory displacement pattern. OSL remained associated with the lowest displacement, OC2.5 and OBL remained among the highest displacements, and OC3 and OC3P remained closely grouped:
\[
d_{\mathrm{OSL}} < d_{\mathrm{ONL}} < d_{\mathrm{OC3}} \approx d_{\mathrm{OC3P}} < d_{\mathrm{OC2.5}} < d_{\mathrm{OBL}}.
\]
This finding supports the internal consistency of the approximation framework across heterogeneous observational probes. However, this six-probe ordering should not be interpreted as a clinical ranking. It belongs to a PCA projection recomputed on the extended dataset and includes mechanically heterogeneous constraints.

The extended analysis therefore reinforces the methodological robustness of the framework without replacing the Level 4 clinical interpretation.

\subsection{Within-session relative positioning of occlusal probes}

The within-session hierarchy provides complementary information to the longitudinal displacement analysis. Whereas the M1--M2 displacement quantifies how much each condition moved over time, the within-session hierarchy describes how each occlusal observational probe was positioned relative to ONL within each measurement session.

At both M1 and M2, ONL, OSL, and OC3 were the three probes closest to the ONL reference. This suggests that OC3 remained close to the natural occlusal reference within the extended latent projection, whereas OC2.5 remained more distant despite being mechanically close to OC3 as a VDO-related probe. This finding reinforces the idea that mechanical proximity does not necessarily imply proximity within the selected PCA representation.

OBL showed the largest change in relative position, being the most distant probe from ONL at M1 but closer to ONL at M2. This observation should be interpreted cautiously because OBL represents open-mouth disengagement without dental contact and is not a therapeutic occlusal position.

The within-session hierarchy should therefore not be interpreted as a clinical ranking. Its role is to describe the relative latent organization of the occlusal probes at each measurement session and to complement the longitudinal viability analysis.

\subsection{Specific interpretation of OBL}

OBL requires a specific interpretation because it is not a VDO-related condition. It corresponds to wide open-mouth disengagement, eliminating dental contacts and modifying the temporomandibular configuration through mandibular opening and condylar translation. Therefore, OBL should not be compared directly with OC2.5, OC3, or OC3P as a therapeutic occlusal configuration.

In the present analysis, OBL was the most distant probe from ONL at M1 and showed the largest longitudinal M1--M2 displacement in the extended six-probe analysis. This indicates that, in the selected latent projection, the open-mouth condition without dental contact was associated with a distinct relative position within the selected PCA projection compared with the ONL reference.

The relative movement of OBL closer to ONL at M2 should not be interpreted as therapeutic improvement. It should be interpreted only as a change in the expression of the system under a heterogeneous open-mouth observational constraint.

\subsection{Held-out and leave-condition-out analyses}

The held-out M2 analysis and the leave-condition-out analysis were included to move beyond simple full-dataset approximation.

In the held-out M2 analysis, the model was evaluated on M2 observations withheld from training within the same single-subject dataset. This analysis showed that the model retained centroid-level approximation ability when tested on unseen observations from the same condition structure. However, pointwise errors remained present, with condition-dependent variability.

In the leave-condition-out analysis, the model was trained on five occlusal observational probes and evaluated on the withheld sixth probe. This is a more demanding test because the model must approximate the transformation of an unseen constraint within the same participant.

The leave-condition-out results showed that approximation was possible for all withheld probes, but with condition-dependent variability. Centroid-level approximation was closest for OC3P, OC3, and OBL, whereas larger deviations were observed for ONL, OC2.5, and OSL. This confirms that leave-condition-out approximation is more demanding than full-dataset approximation and should not be interpreted as clinical prediction.

These internal tests do not establish generalizable predictive validity. They do, however, strengthen the Level 5 claim by showing that the model is not limited to a single full-dataset fit. It can approximate held-out observations and, to a more limited extent, withheld observational probes within the same subject.

\subsection{Occlusal probes as observational constraints}

A central feature of the proposed framework is the distinction between longitudinal system transformation and observational constraint. The occlusal configurations were applied only during gait acquisition and were not imposed continuously between M1 and M2. They should therefore be interpreted as observational probes revealing the expression of the system under constraint at discrete time points, not as continuous causal drivers of longitudinal evolution.

The M1--M2 transformation should instead be understood as a combined longitudinal context. It includes intrinsic system evolution, structured sensorimotor intervention, spontaneous variability, daily-life conditions, and other contextual factors that cannot be separated in the present design.

Consequently, the model does not isolate the independent effects of occlusion, intervention, time, or spontaneous evolution. It approximates the combined transformation observed between M1 and M2 in the representation space.

This distinction prevents a causal misinterpretation of the results. The present data do not show that an occlusal condition caused a specific gait evolution. They show that the system, when observed under different occlusal probes at M1 and M2, displayed condition-dependent representational displacement patterns that could be internally approximated.

\subsection{Predictive approximation and representation learning}

The predictive formulation introduced here connects the clinical multi-level framework with predictive representation-space approaches in representation learning \cite{bengio2013,lecun2022}. The shared principle is that system evolution may be represented directly in an embedding space rather than reconstructed at the level of high-dimensional observable variables.

In this sense, the present approach is conceptually related to joint-embedding predictive architectures. The model does not attempt to reconstruct the original high-dimensional gait variables. Instead, it approximates the future latent position of the system directly within the PC1--PC2 representation. The prediction is therefore performed in representation space, from an observed M1 representation toward an approximated M2 representation.

However, the present implementation should not be interpreted as a formal JEPA model. It does not use separate learned encoders for input and target states, does not implement a dedicated joint-embedding architecture, and does not aim to learn a general world model. It remains a simplified supervised neural-network approximation of one observed longitudinal transition in a single participant.

The conceptual decomposition between \(G_\theta\), representing longitudinal transformation, and \(H_\phi\), representing observational constraint, provides a useful formal structure. However, this decomposition was not empirically identifiable in the present implementation. The simplified neural network approximated the combined effect of transformation and observational constraint rather than estimating them as two distinct modules.

This distinction is essential. The mathematical decomposition structures the reasoning, but the available data do not allow separate estimation of its components. The value of the model is therefore not architectural generality, but the demonstration that the observed viability-related transformation can be approximated internally in a latent representational space.

\subsection{Toward internal approximation of longitudinal displacement}

Within the present framework, Level~5 concerns the approximation of an observed longitudinal descriptor rather than prediction of clinical viability.

At Level~4, longitudinal behavior was summarized retrospectively through centroid displacement in a selected PCA representation. At Level~5, the same representation-dependent displacement is approximated by a supervised model.

This shift does not establish a viability biomarker and does not allow prediction of whether an occlusal configuration will remain clinically stable or functional. It provides only a methodological demonstration that the observed displacement pattern can be internally approximated in one subject.

The present findings therefore support the restricted progression
\[
\text{static representational non-identifiability}
\rightarrow
\text{observed longitudinal centroid displacement}
\rightarrow
\text{internal approximation of the observed displacement}.
\]

This progression remains exploratory. It does not establish causal occlusal effects, validated dynamical thresholds, therapeutic superiority, or generalizable prediction.

\subsection{Transition toward Level 6}

The present Level 5 analysis should be understood as the last single-subject step before multi-walker predictive modeling. It tests whether viability-related represented trajectories can be internally approximated before asking, at Level 6, whether such trajectories can be predicted across individuals.

A future Level 6 framework would require multiple walkers, repeated longitudinal transitions, and prospective evaluation. In such a framework, the model would no longer be trained and evaluated only within one subject. Instead, it would need to learn from several individuals and predict the viability-related trajectory of a new walker or a new clinical configuration.

Only under such conditions could predictive viability move from internal approximation toward clinically operational modeling.

\section{Limitations}

Several limitations must be explicitly acknowledged.

First, the analysis is based on a single-subject dataset. The observed patterns may reflect subject-specific dynamics and cannot be generalized to a Parkinsonian population, to other neurological conditions, or to other clinical contexts.

Second, the predictive model was trained and evaluated within the same single-subject dataset. Although held-out M2 and leave-condition-out analyses provide stronger internal tests than a full-dataset fit alone, they do not establish generalizable predictive validity.

Third, the predictive model is intentionally simplified. The neural network was used only to test whether the observed coordinate transformation could be approximated. It does not capture the full complexity of biomechanical, neurological, occlusal, or sensorimotor processes.

Fourth, the representation relies on PCA. PCA provides a reproducible linear coordinate system, but the first two components retain only part of the total variance. Reported distances are therefore distances within the selected PC1--PC2 plane, not complete measurements of physiological or whole-system dynamics.

Fifth, the PCA projection differs between the core Level~4-aligned analysis and the extended six-probe analysis. Consequently, numerical distances from the two projections should not be compared directly.

Sixth, the transformation observed between M1 and M2 reflects a combined longitudinal context including intrinsic evolution, structured sensorimotor intervention, spontaneous variability, daily-life conditions, and unmeasured factors. These components cannot be disentangled.

Seventh, occlusal conditions were applied only during measurement and not continuously between M1 and M2. Their role is limited to probing the expression of the participant within the selected representation. They cannot be interpreted as causal drivers of longitudinal evolution.

Eighth, the training pairs do not represent physiologically matched stride-level correspondences. They were constructed computationally by aligning observations within each condition after truncation to the minimum available number of observations. This supports condition-level approximation but not stride-by-stride prediction.

Ninth, the descriptor-based encoding of occlusal probes was introduced to permit leave-condition-out testing. It is a pragmatic modeling choice and not a validated biomechanical parameterization of occlusal state.

Finally, viability is not directly estimated in the present study. The model approximates a representation-dependent displacement proxy inherited from Level~4. The viability region $\mathcal{V}$ remains conceptual and does not constitute a validated physiological or clinical construct.

Taken together, these limitations define the present work as a Level~5 proof-of-concept supporting internal approximation of an observed representation-dependent transformation, not generalizable clinical prediction.

\section{Conclusion}

This study introduces a Level~5 centered on internal approximation of observed longitudinal centroid displacement within a single-subject gait dataset.

The revised Level~3 study showed static representational non-identifiability rather than clearly separated condition-specific states. Level~4 subsequently described condition-dependent M1--M2 displacement within a common PCA representation. The present study examined whether this observed coordinate transformation could be approximated by a simplified supervised model.

In the core Level~4-aligned analysis, the model preserved the Euclidean displacement hierarchy:
\[
d_{\mathrm{OC3}} < d_{\mathrm{ONL}} < d_{\mathrm{OC2.5}}.
\]

In the extended six-probe analysis, the model preserved the broad structure of the exploratory displacement ordering. Held-out M2 and leave-condition-out analyses showed that internal approximation remained possible beyond the full-dataset fit, but with condition-dependent error.

The within-session analysis described relative centroid positions with respect to ONL. It did not identify distinct physiological states, latent functional equivalence, or a therapeutic ranking.

These findings do not establish causal occlusal effects, clinical predictive validity, validated viability thresholds, therapeutic superiority, or generalizable patient-level forecasting. They show only that an observed, representation-dependent longitudinal transformation can be internally approximated within the same single-subject dataset.

The contribution of Level~5 is therefore methodological. It provides a computational bridge from retrospective description of centroid displacement toward future multi-subject modeling. A Level~6 framework would require multiple walkers, repeated longitudinal transitions, prospective validation, fixed shared representations, and explicit evaluation on unseen individuals.

\section*{Author Contributions}

Jacques Raynal: conceptualization, study design, data analysis, supervised machine-learning implementation, predictive modeling, interpretation, and writing-original draft.

Pierre Slangen: biomechanical interpretation, methodological review, and writing-review and editing.

Elsa Raynal: delivery of the sophrology-oriented sensorimotor intervention, description of the intervention context, contribution to the clinical interpretation of sensorimotor regulation, and writing-review and editing.

Jacques Margerit: conceptual supervision, clinical and scientific interpretation, and writing-review and editing.

All authors reviewed and approved the final manuscript.

\section*{Data and Code Availability}

The raw clinical gait recordings are not publicly distributed in the present preprint because they derive from a single clinical participant. Anonymized derived datasets sufficient to reproduce the numerical analyses, together with the complete Python script used for preprocessing, PCA projection, centroid computation, supervised machine-learning approximation, held-out M2 evaluation, leave-condition-out evaluation, within-session hierarchy analysis, and export of derived tables, may be made available by the corresponding author upon reasonable request.

The Python script was designed to reproduce the numerical results reported in the manuscript from structured CSV input files. It does not contain personally identifiable information.
\section*{Ethical Considerations}

The participant provided written informed consent for the use of anonymized gait data for research and publication. No personally identifiable information is reported.


\end{document}